\documentclass{bioinfo}
\copyrightyear{2015} \pubyear{2015}
\usepackage{multirow}

\access{Advance Access Publication Date: Day Month Year}
\appnotes{Manuscript Category}

\begin{document}
\firstpage{1}
\subtitle{Subject Section}

\title[short Title]{BridgeDPI: A Novel Graph Neural Network for Predicting Drug-Protein Interactions}
\author[Sample \textit{et~al}.]{Yifan Wu\,$^{\text{\sfb 1, 2, \dag}}$, Min Gao\,$^{\text{\sfb 1, \dag}}$, Min Zeng\,$^{\text{2,}}$, Feiyang Chen\,$^{\text{\sfb 1}}$, Min Li\,$^{\text{\sfb 2,}*}$ and Jie Zhang\,$^{\text{\sfb 1,3,}*}$}
\address{$^{\text{\sf 1}}$SenseTime Research, Shanghai, 200233, China,\\
$^{\text{\sf 2}}$School of Computer Science and Engineering, Central South University, Changsha, 410083, China,\\
$^{\text{\sf 3}}$Qing yuan Research Institute, Shanghai Jiao Tong University, Shanghai, China \\
$ $\\
$^{\text{\sf \dag}}$These authors contributed equally to this work. }

\corresp{$^\ast$To whom correspondence should be addressed.}

\history{}

\editor{}

\abstract{\textbf{Motivation:} Exploring drug-protein interactions (DPIs) work as a pivotal step in drug discovery. The fast expansion of available biological data enables computational methods effectively assist in experimental methods. Among them, deep learning methods extract features only from basic characteristics, such as protein sequences, molecule structures. Others achieve significant improvement by learning from not only sequences/molecules but the protein-protein and drug-drug associations (PPAs and DDAs). The PPAs and DDAs are generally obtained by using computational methods. However, existing computational methods have some limitations, resulting in low-quality PPAs and DDAs that hamper the prediction performance. Therefore, we hope to develop a novel supervised learning method to learn the PPAs and DDAs effectively and thereby improve the prediction performance of the specific task of DPI. \\
\textbf{Results:} In this research, we propose a novel deep learning framework, namely BridgeDPI. BridgeDPI introduces a class of nodes named hyper-nodes, which bridge different proteins/drugs to work as PPAs and DDAs. The hyper-nodes can be supervised learned for the specific task of DPI since the whole process is an end-to-end learning. Consequently, such a model would improve prediction performance of DPI. In three real-world datasets, we further demonstrate that BridgeDPI outperforms state-of-the-art methods, achieving AUC of 0.989 (for seen proteins), 0.952 (for unseen proteins) in the customized BindingDB dataset, 0.995 in the C.elegans dataset, 0.990 in the human dataset. Moreover, ablation studies verify the effectiveness of the hyper-nodes. Last, in an independent verification, BridgeDPI explores the candidate bindings among COVID-19's proteins and various antiviral drugs. And the predictive results accord with the statement of the World Health Organization and Food and Drug Administration, showing the validity and reliability of BridgeDPI. Hopefully, the BridgeDPI method will facilitate and accelerate real-world drug discovery and drug screening. \\
\textbf{Availability:} The source code of BridgeDTI can be accessed at https://github.com/DeepAAI/BridgeDPI.\\
\textbf{Contact:} \href{name@bio.com}{limin@mail.csu.edu.cn or zhangjie1@sensetime.com}
}

\maketitle

\section{Introduction}

\begin{figure*}
    \centering
    \includegraphics[scale=0.4]{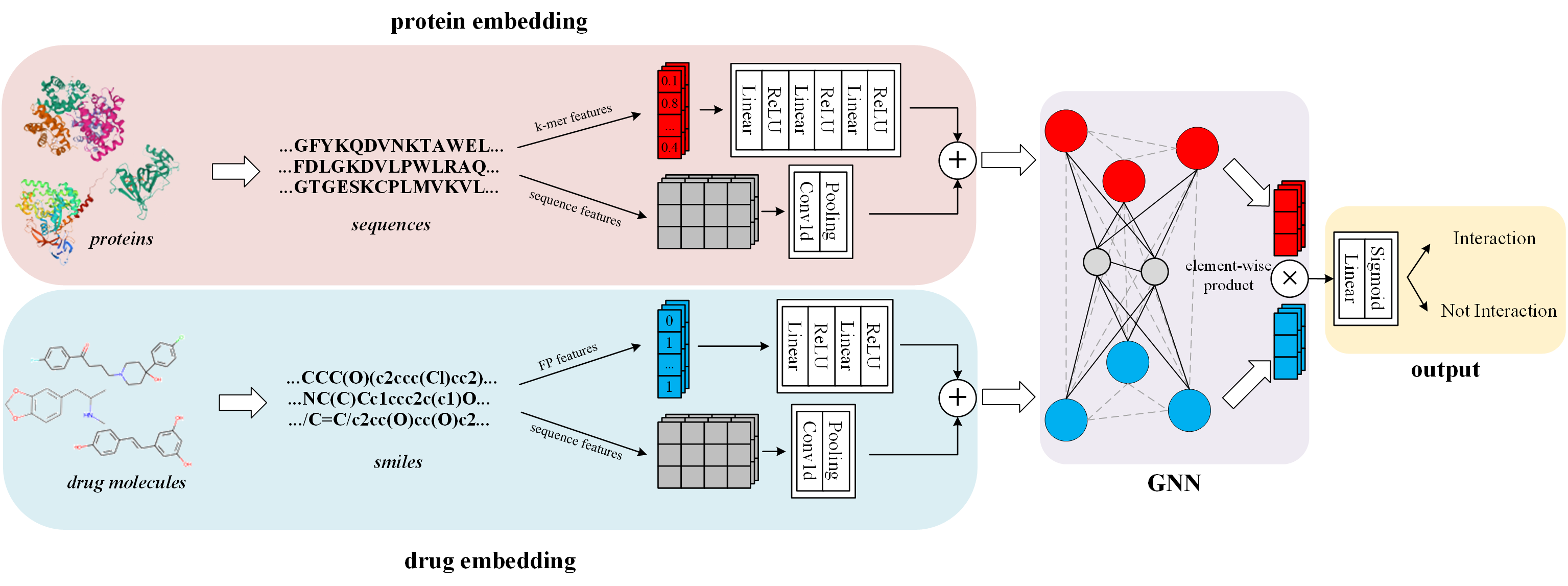}
    \caption{ Model architecture: protein sequences and drug molecules are the inputs of BridgeDPI; first, the k-mer/sequence features of proteins and FP/sequence features of drugs are obtained. Then through the multilayer perceptron layers, the node embeddings are calculated and fed to the GNN; Finally, the outputs of GNN are multiplied and a full connected layer with sigmoid activation is applied. }
    \label{fig:01}
\end{figure*}

The drug discovery and drug screening are complex. The typical timeline usually takes 10-20 years and costs US\$0.5-2.6 billion \citep{avorn20152,paul2010improve}. Among them, exploring possible drug-protein interactions (DPIs) is a crucial step. Although experimental assays remain the most reliable approach for determining DPIs, they are time-consuming and cost-intensive. Therefore, efficient computational methods for predicting protein-drug interactions are significant and urgently demanded.

Current DPI prediction methods can be summarized as three forms: docking-based methods, machine learning-based methods, and deep learning-based methods. Docking-based methods look for the best binding position inside the binding pocket of the proteins for drug molecules \citep{2015Molecular,2017Protein}. However, it takes a lot of time and lacks available 3D protein structures for a large-scale dataset. Machine learning-based methods \citep{ml1,ml2,ml5} usually use handcrafted features. However, one needs to choose, combine and compare these handcrafted features carefully, which require expertise and experience. These methods usually use handcrafted features before modeling, which require much expertise. Recently, with the plentiful accumulation of data, deep learning-based methods have been successfully applied to various bioinformatics tasks \citep{2019Protein,2020KAICD,min2017deep,zhang2019deepfunc,zhang2020deep}. It has further improved the performance of DPI prediction by using the deep structure and vast learnable parameters. Deep learning models such as DeepDTA \citep{deepdta}, WideDTA \citep{widedta}, DeepConv-DTI \citep{DeepConv-DTI}, PADME \citep{padme}, GraphDTA \citep{nguyen2019graphdta}, E2E \citep{ibm}, drugVQA \citep{zheng2020predicting} have similar steps: 1) encode proteins and drugs. 2) design a feature extractor modules to capture high-level features of proteins and drugs. 3) fuse the high-level features of proteins and drugs, and perform prediction through full connected layers. The disadvantage of these methods is that they neglect to leverage the protein-protein associations (PPAs) and drug-drug associations (DDAs). The magic of PPAs and DDAs comes from the fact that proteins usually interact with similar drugs \citep{2011An,Ana2017Network}. Therefore, the information in DDAs and PPAs would improve the DPI prediction. A large number of conventional methods that used PPAs and DDAs convinced the effectiveness of DDAs and PPAs in the DPI prediction \citep{chen2012drug}. 

Generating PPAs and DDAs mainly has three types of methods: structure-based methods, sequence-based methods and the methods based on known DTIs. But there are still some limitations. For structure-based methods, although they can measure PPAs accurately, they are limited by the availability of protein structural data. For sequence-based methods, PPAs are generally computed by using BLAST which is based on multi sequence alignment and homology information. The results of BLAST depend on the scale of protein data. When researchers only have hundreds or thousands of protein sequences in their study, using BLAST on such a small number of protein sequences cannot find their homological proteins for most of the sequences. Thus these sequences cannot provide useful information to represent PPAs appropriately. For the methods based on known DTIs, it uses negative sampling to generate negative samples for PPAs and DDAs without any reliable biological evidence. However, negative sampling treat the unknown drug-protein pairs as negative. However, the negative drug-protein pairs could be positive. 




To tackle the above limitations and further improve the prediction performance, we develop BridgeDPI, a deep learning framework for predicting DPIs. The novelty of BridgeDPI is that we introduce the hyper-nodes to connect different proteins/drugs. The role of hyper-nodes is to construct bridges between proteins/drugs. The bridges implicitly measure the associations among proteins/drugs and therefore can be used as the networks of PPAs and DDAs. On one side, the hyper-nodes are automatically learned for DPI prediction. The quality of the learned PPAs and DDAs is guaranteed by the back propergation of the end-to-end learning of BridgeDPI. On the other side, some unknown biological connections may also be explored by the hyper-nodes. In brief, the hyper-nodes are considered as the essential elements to improve the prediction performance. 

We demonstrate the superior performance of BridgeDPI: the AUC scores of 0.989 (for seen proteins), 0.952 (for unseen proteins) in the customized BindingDB dataset, 0.995 in the C.elegans dataset, 0.990 in the human dataset.

\begin{methods}
\section{Methods}
\subsection{Overview}
The overview of BridgeDPI is shown in Figure \ref{fig:01}. BridgeDPI contains two inputs: the protein sequence and the drug smiles. For protein sequences, a three-layer feed-forward network is used to process the k-mer features. In addition, a Convolutional Neural Network (CNN) is used to extract the sequence features of proteins. We combine the two features to obtain the final protein embeddings. For drug smiles, a two-layer feed-forward network is used to process the molecular fingerprint features. In addition, a CNN is used to extract the sequence features of drugs. We also combine the two features to obtain the final drug embeddings. After that, we introduce some hyper-nodes to construct the bridges between proteins/drugs and use Graph Neural Network (GNN) to learn the associations between proteins/drugs. Finally, we get element-wise products of the protein and drug outputs after GNN, and then use a linear layer with sigmoid activation to predict the interactions.

\subsection{Embedding of proteins and drugs}

Before feeding protein sequences and drug molecules to BridgeDTI, they need to be encoded as numeric vectors. For a long time in the past, the k-mer and FingerPrint (FP) features have been very effective to represent proteins and drugs. Since the advent of deep learning techniques, people begin to characterize proteins and drugs at the amino acid level and atomic level by using CNN, GNN, etc. However, in fact, k-mer and FP features cannot be replaced completely by the sequence or graph features. Thus, we reintroduce k-mer and FP features, which are effective but ignored by many researchers at present.

For proteins, k-mer is a classical and effective method for protein embedding. K-mer features can describe the type and number of amino acid functional groups, which are very important in the prediction of DPIs. As a result, k-mer features are used to represent the proteins. In our study, we set $k$ as 1,2,3, and thus we got $20^1+20^2+20^3=8420$ k-mer features.
\begin{equation}
x^{i}=(x^i_1,x^i_2,...,x^i_{8420})\label{eq:01}\vspace*{-0pt}
\end{equation}
where $x^i_{1...20}, x^i_{21...420}, x^i_{421...8420}$ represent 1-mer, 2-mer, 3-mer features of protein $i$, respectively. We normalize each part horizontally to eliminate the effect of protein length, as shown in Formula~ (\ref{eq:02}).  
\begin{equation}
    \mathrm{normalized}(x^{i}_{1...20})=\frac{x^{i}_{1...20}-\mathrm{mean}(x^{i}_{1...20})}{\mathrm{std}(x^{i}_{1...20})}\label{eq:02}\vspace*{-0pt}
\end{equation}
where $x^{i}_{1...20}$ represents 1-mer feature of protein $i$, $\mathrm{mean}(\cdot)$ and $\mathrm{std}(\cdot)$ are to calculate the mean and standard deviation. We also do the same for the 2-mer part $x^i_{21...420}$ and 3-mer part $x^i_{421...8420}$.

For drug molecules, FP is a very efficient technique in drug discovery and virtual screening. In molecular FP, the topological information of molecular structure is encoded as a vector, which can well represent a drug molecule. Consequently, we use Morgan FP \citep{fingerprint} to represent a molecule. It can encode a drug as a vector with dimensions of 1024.

\begin{equation}
    F^{j}=(b_{1},b_{2},{\ldots},c_{1024})\label{eq:03}\vspace*{-0pt}
\end{equation}
where $F^{j}$ represents FP features of drug $j$, $b_{1...1024}$ are the binary values in the fingerprint. 

After getting the characterization of proteins $A^{i}$ and drugs $F^{j}$, their relationship cannot be constructed directly. Because the two vectors have different dimensions and do not belong to the same space. Therefore, they need to be mapped into the same space through several layers of neural networks. At the same time, we use CNN with max-pooling to extract the sequence features of proteins and drugs, respectively, and then fuse them together.

\begin{gather}
    u^{i}=\mathrm{f}_{p}(A^i)+p^{i} \\ v^{j}=\mathrm{f}_{d}(F^j)+d^{j}\label{eq:04}
\end{gather}
where $f_{p}(\cdot)$ and $f_{d}(\cdot)$ are the nonlinear transformation layers for protein and drug, respectively, $p^{i}$ and $d^{j}$ are the sequence features extracted by CNNs. Finally, $u^{i}$ is used as the embedding of protein $i$. $v^{j}$ is used as the embedding of drug $j$.

\subsection{Introduction of hyper-nodes}

After getting the embedding of proteins and drugs, how can we predict the interactions between proteins and drugs, especially those unconnected protein-drug pairs candidates? It is the main challenge to DPI predictions due to the lack of sufficient neighborhood information. As shown in Figure \ref{fig:02}, given known interacted pairs, we need to infer unknown relationship between protein $p_2$ and drug $d_2$. Traditional methods usually computed DDAs and PPAs to construct a heterogeneous network. And then according to the associations between proteins/drugs, we can construct a path $p_2-p_1-d_1-d_2$ to infer the relationship. However, as mentioned above, it is difficult to acquire high-quality PPAs and DDAs. Therefore, we introduce a new kind of virtual nodes, namely hyper-nodes. These nodes connect all proteins and drugs, which are randomly initialized and updated during the training process.

With these hyper-nodes, we can construct some bridges between proteins/drugs and capture the relationship between them. It makes our model learn a higher-quality PPA and DDA information from the data and further improve the predictive performance of DTI. As illustrated in Figure \ref{fig:02}, although we do not know the relationship between $p_1$ and $p_2$, by introducing a hyper-node $h_1$, we can get the relationship between $p_1$ and $h_1$, $p_2$ and $h_1$. Therefore, the relationship between $p_1$ and $p_2$ can be derived from $p_2-h_1-p_1$. In the same way, the relationship between $d_1$ and $d_2$ can be derived from $d_2-h_2-d_1$. Finally, the relationship between $p_2$ and $d_2$ can be inferred through $p_2-h_1-p_1-d_1-h_2-d_2$. 

\begin{figure}
    \centering
    \includegraphics[scale=0.3]{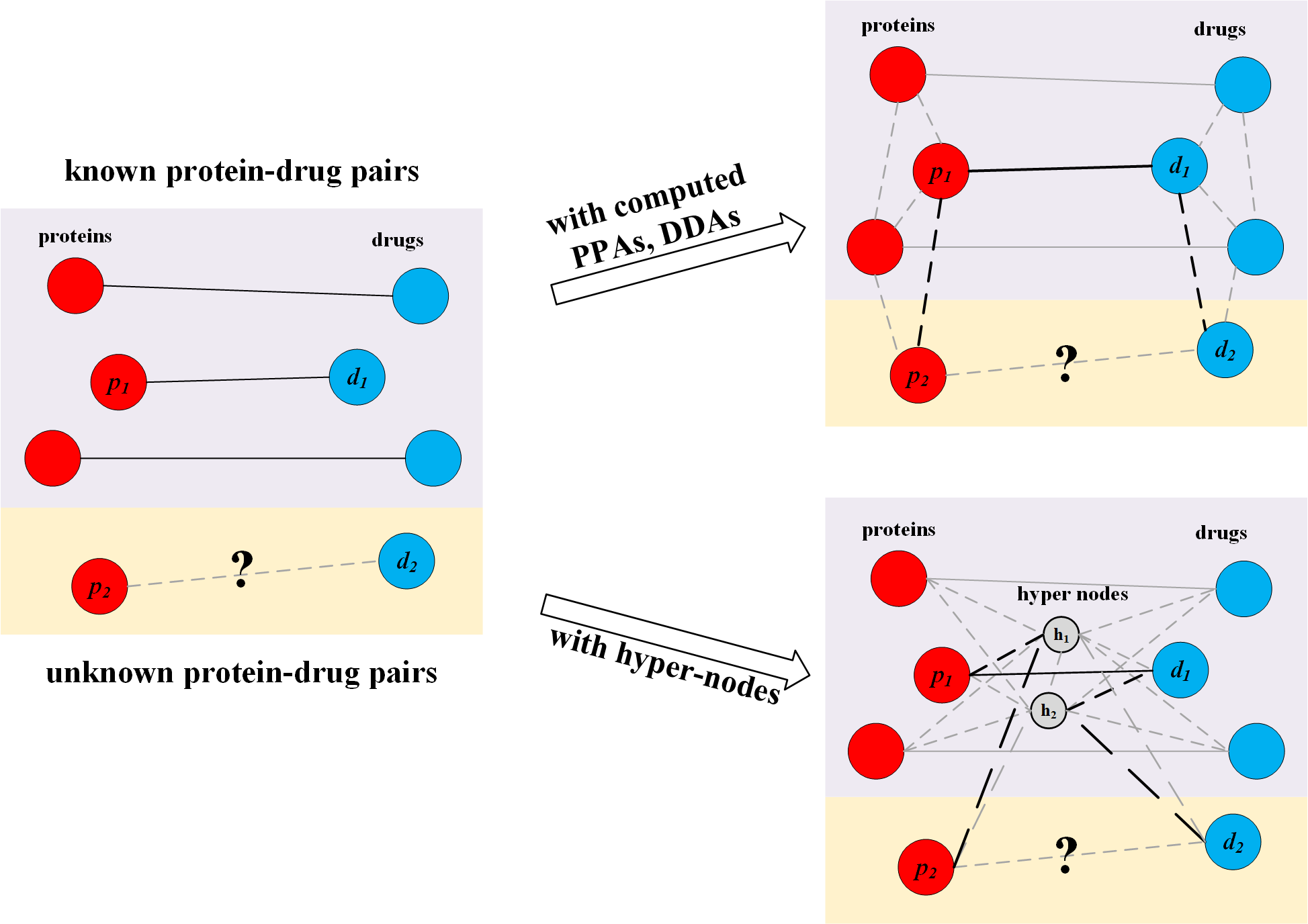}
    \caption{\centering Comparison of the effects of introducing hyper-nodes and using computed DDA, PPA information}
    \label{fig:02}
\end{figure}

\subsection{Graph neural network}

Assume that the set of hyper-nodes is $\{n_{1},n_{2},\ldots,n_{m}\}$, where $m$ is the number of hyper-nodes, and hyper-nodes have the same embedding size with $u^{i}$ and $v^{j}$. For each protein-drug pair in dataset, we can get a graph including nodes ${u^{i},v^{j},n_{1},n_{2},\ldots,n_{m}}$ with edges calculated by cosine similarity.

\begin{equation}
A_{i,j}=\frac{n_{i}\cdot n_{j}}{\lVert n_{i}\rVert_{2} \lVert n_{j}\rVert_{2}}\label{eq:05}
\end{equation}
where $n_{i}$ and $n_{j}$ represent any two nodes in the graph (including $u$ and $v$), $\lVert \cdot \rVert_2$ is to compute 2-norm of a vector. In fact, $A_{i,j}$ could be negative, which are not suitable for the calculation of GNN. Thus, we need to filter out the negative edges by $\mathrm{ReLU}(\cdot)$, which means there’s no edge if cosine similarity is less than zero. And the computational steps of GNN are as follows: 

\begin{gather}
L = D^{-1/2}\mathrm{ReLU}(A)D^{-1/2} \label{eq:06}\\
Z^{0} = (u^{i},v^{j},n_{1},n_{2},\ldots,n_{m})^\mathrm{T}\label{eq:07} \\
Z^{i} = \mathrm{ReLU}(LZ^{(i-1)}W^{i}+b^{i})+Z^{i-1}\label{eq:08} 
\end{gather}
where $A \in \mathbb{R}^{(m+2)\times(m+2))}$ is the similarity matrix, $Z^0 \in \mathbb{R}^{(m+2)\times(d)}$ is the embedding matrix of nodes, $d$ is the embedding size of nodes, $D \in \mathbb{R}^{(m+2)\times(m+2)}$ is the degree matrix, $Z^{i}\in \mathbb{R} ^{(m+2)\times d)}$ is the output of $i$-th layer of GNN, $W^{i}$ and $b^{i}$ are the parameters in $i$-th layer of GNN. Additionally, residual structure \citep{he2016deep} is applied in each layer of GNN, as shown in Equation~(\ref{eq:08}). It is a very effective structure in deep networks because it guarantees that some shallower network may exist in the model. 

\subsection{Prediction of protein-drug interactions}
The output of GNN belongs to $\mathbb{R}^{(m+2)\times d}$, and the first two rows are the embeddings of protein nodes and drug nodes after fusing graph information. The two embeddings are taken out to do element-wise multiplication and then fed to a fully connected layer. Finally, the $\mathrm{sigmoid}(\cdot)$ is applied to classify whether the protein-drug pairs has interactions. 

\begin{gather}
h_{ij}=\hat{u}^{i}\otimes\hat{v}^{j} \label{eq:09} \\
\tilde{y}_{ij}=\mathrm{sigmoid}(f_{o}(h_{ij}))\label{eq:10}
\end{gather}
where $\hat{u}^{i}$ and $\hat{v}^{j}$ are the first two rows of the final output matrix in GNN, $f_{o}(\cdot)$ is the fully connected layer, $\tilde{y}_{ij}$ represents the probability of the interaction between protein $i$ and drug $j$, $\otimes$ means element-wise product. 

\subsection{Other details}
Batch normalization \citep{ioffe2015batch}, which standardizes each batch of data, is a very effective method to accelerate training and improve the performance of the deep network. Dropout \citep{srivastava2014dropout}, which randomly drops part of neuron units, can greatly improve the generalization of the model. We add batch normalization and dropout after each layer in the model. Finally, the model is trained by minimizing the cross-entropy loss function.

\begin{equation}
L(y_{ij},\tilde{y}_{ij}\mid \theta)=-(y_{ij}log\tilde{y}_{ij}+(1-y_{ij})log(1-\tilde{y}_{ij})+\lambda \sum\nolimits_{\theta \in\Theta}\|\theta\|^{2} \label{eq:11}
\end{equation}
where $\theta$ is the set of all model parameters, $\lambda$ is the regularization coefficient. 

\end{methods}

\section{Experiments}

\begin{figure*}[h]
    \centering
    \includegraphics[trim=50 0 50 0, clip,scale=0.65]{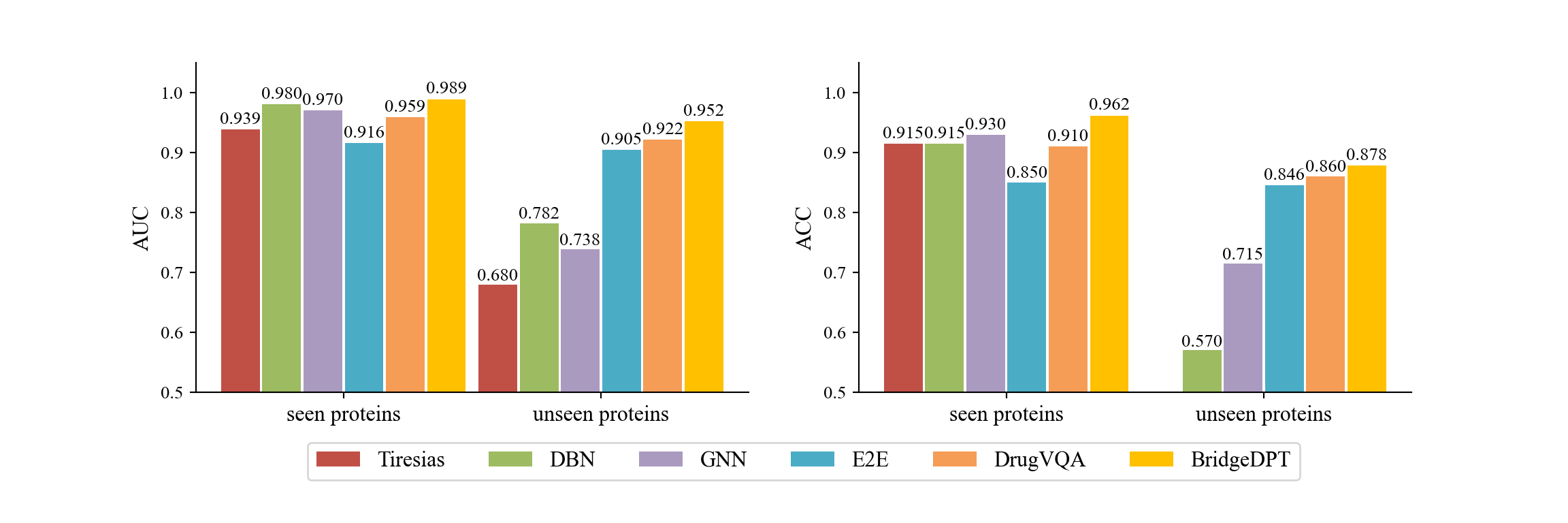}
    \caption{Comparison of BridgeDPI with five baselines: (left) shows AUC for seen proteins and unseen proteins in the test; (right) shows ACC for seen proteins and unseen proteins in the test. Note that the accuracy scores of Tiresias do not show in (right) because they are lower than the lower bound of the y-axis. }
    \label{fig:04}
\end{figure*}

\subsection{Datasets}

\noindent\textbf{BindingDB dataset.} BindingDB is a public, web-accessible database for measuring binding affinities, focusing on interactions between proteins and small drug molecules. It contains a total of 2,067,981 binding data, for 8,161 protein targets and 910,476 small molecules \citep{gilson2016bindingdb}. Based on the raw BindingDB dataset, Gao et al. constructed a binary classification dataset containing 39,747 positive samples and 31,218 negative samples \citep{ibm}. A protein-drug sample is positive if IC50 is less than 100nm, or negative if IC50 is greater than 10,000nm. They divided the data into training set (28,240 positive samples, 21,915 negative samples), validation set (2,831 positive samples, 2,776 negative samples), and test set (2,706 positive samples, 2,802 negative samples) with a guaranteed ratio of positive to negative samples. We choose this customized BindingDB as one of our benchmark datasets for head-to-head comparisons.

\noindent\textbf{C.elegans and human dataset.} Liu et al. obtained a set of highly credible negative samples of DPI via silico screening method \citep{liu2015improving}. Combined with the known positive samples, they constructed two dataset C.elegans and human respectively. Following Tsubaki et al. \citep{tsubaki2019compound}’s work, we choose the balanced versions of these datasets. For the C.elegans dataset, there are 7,786 binding samples (3,893 positive samples, 3893 negative samples), with 1,876 protein targets and 1,767 drug molecules. For the human dataset, 6,728 binding data are included, with 2,001 protein targets and 2,726 drug molecules. These two datasets are randomly divided into 5 folds respectively, and each time one fold is selected as the validation set for 5-fold cross-validation.

\noindent\textbf{DUD-E dataset.} DUD-E is a widely used dataset covering 102 proteins and 22,886 clustered ligands \citep{mysinger2012directory}. There are 50 decoys for each activity with similar physical and chemical properties but dissimilar 2D topology. It contains 1,429,790 protein-ligand samples in total (22,645 positive samples, 1,407,145 negative samples). Following Zheng et al. \citep{zheng2020predicting}’s work, we perform 3-fold cross-validation in this dataset and the average evaluation metrics are reported. In addition, DUD-E is used as an independent test set to evaluate the performance of models in the real world.

\subsection{Training details}
The proposed model is implemented with Pytorch 1.6.0 \citep{paszke2019pytorch} and the parameters are initialized by default. We use Adam \citep{kingma2019method} optimizer with a learning rate of 0.001 to adjust the parameters in the training process. In order to prevent over-fitting, L2 regularization is added to the loss function. For each step, a batch of protein-drug pairs is randomly selected to run the gradient descent algorithm and the batch size is set to 512. For the setting of other hyper-parameters, such as the number of layers, the number of neurons, the ratio of dropout, etc., many experiments are carried out to choose the values according to the performances on the validation set. The maximum number of epochs is set to 100, and the models with the best area under ROC curve AUC in the validation set are saved. Finally, the saved models are evaluated on a test set by metrics of accuracy (ACC) and the AUC. Besides, our model has low computational complexity, high parallelism, and fast training speed. The training process of the customized BindingDB dataset will be finished in about 15 minutes on the GPU platform of GTX 1080ti.

For proteins, we introduce a 2-layers network for nonlinear transformation, containing 1024, 128 neurons, respectively. For drug molecules, we introduce a 3-layers network for nonlinear transformation, containing 1024, 256, 128 neurons, respectively. The outputs are 128-dimensional vectors that serve as the input of node embeddings on the complete DPI graph. Then, we use a 3-layers GNN, which means that each node has aggregated three layers of neighbor information. In the end, the scores of the interactions between proteins and drugs are obtained through a 2-layers feed forward network. Moreover, we dropout 50\% neurons after each layer. And the L2 regularization coefficient is set to 0.001 to limit the model ability and prevent over-fitting.

\subsection{Results}

First of all, we conduct experiments on the customized bindingDB dataset which is extracted from the bindingDB by Gao et al. Following their settings, we also make the same division of the dataset to ensure that the validation set and test set contain some unseen proteins (that are not appear in the training set), which is closer to the real world. Based on the prediction results in the test set, we calculate AUC and ACC (threshold is 0.5, same as below) of BridgeDPI with the best parameters. To compare the performance with other methods, we choose Tiresias \citep{fokoue2016predicting}, DBN \citep{wen2017deep}, GNN \citep{tsubaki2019compound}, E2E \citep{ibm} and DrugVQA \citep{zheng2020predicting} as baselines. As we know, there are a large number of unknown proteins in nature, and this is why we should focus on predicting the new proteins, i.e., cold-start problem. Therefore, the test set is divided into an unseen protein set (the proteins that do not appear in the training set) and a seen protein set (the proteins that appear in the training set). As shown in Figure \ref{fig:04}, the existing models generally achieve good performances in the seen protein set (AUC exceeds 0.9, ACC exceeds 0.85), but differ greatly in the unseen protein set. Based on a traditional relational network, Tiresias performs poorly on the unseen proteins, with an AUC of only 0.68. Tiresias is a method based on computed similarity information to obtain the features of proteins and drugs, and it predicts DPIs through a logistic regression model. We hypothesize that only based on a linear model using similarity information, the expression of the unseen proteins may not be sufficient, resulting in the ACC on unseen proteins is even less than 0.5. DBN, E2E, DrugVQA, and BridgeDPI have AUC over 0.9 and ACC over 0.8 in unseen proteins, indicating the effectiveness of deep learning techniques in solving unseen proteins. Among them, our BridgeDPI outperforms other baselines and achieves start-of-the-art performances, with AUC and ACC reaching 0.987 and 0.954 in seen proteins, 0.951 and 0.887 in unseen proteins. It indicates that the introduction of hyper-nodes indeed improves the expression of proteins/drug features, and the deep graph neural network also enables BridgeDPI to learn the deeper interaction rules between proteins and drugs.  

Furthermore, we also conduct experiments on the C.elegans dataset and human dataset \citep{liu2015improving} which are widely used in many studies. The results are shown in Table \ref{tab:02}. Since Gao et al. \citep{ibm} do not provide the code of E2E, we reproduced the model and obtain experimental results on the two datasets without using the Gene Ontology (GO) features. Other results of baselines are from their original papers \citep{tsubaki2019compound,zheng2020predicting}. From Table \ref{tab:02}, for randomly divided C.elegans and human datasets, almost all proteins in the test set are seen, which means models can learn all protein information better from the training dataset, resulting in very good results. In this case, the unsupervised k-NN is slightly worse than other models, with AUC 0.858 and F1 0.814 on the C.legans dataset, AUC 0.860 and F1 0.858 on the human dataset, respectively. The supervised machine learning methods (RF, L2, SVM) are slightly better, with AUC of the C.elegans dataset reaching around 0.9, AUC of the human dataset exceeding 0.9. In contrast, GNN, E2E/GO, DrugVQA, and BridgeDPI based on deep learning methods perform very well, with AUCs over 0.97 and F1s over 0.9. Among them, BridgeDPI achieves the best performances, with AUC, precision, recall, and F1 of 0.995, 0.980, 0.965, 0.972 on the C.elegans dataset, respectively, and 0.990, 0.963, 0.949, 0.956 on the human dataset, respectively. The results are in line with our expectations. Because the models such as KNN, RF, L2, and SVM, without high-quality features, are difficult to learn complex nonlinear relationships (protein-drug interaction), while the deep learning models have strong feature extraction abilities to learn the interaction rules. On this basis, BridgeDPI integrates PPA and DDA information, further improving the results.

\begin{table}[h]
\centering
\caption{ \centering Performances of BridgeDPI with other models on different dataset$^1$ } 
\setlength{\tabcolsep}{2mm}{
\begin{tabular}{c|c|c|c|c}
\hline
Models & AUC & Precision & Recall & F1  \\
\hline
\multicolumn{5}{c}{C.elegans dataset} \\
\hline
k-NN & 0.858 &0.801  &0.827  &0.814 \\
RF & 0.902& 0.821&  0.844&  0.832\\
L2 & 0.892& 0.890&  0.877&  0.883\\
SVM & 0.894 &0.785& 0.818&  0.801\\
GNN & 0.978&0.938&  0.929&  0.933\\
E2E/GO & 0.986& 0.950& 0.950& 0.950\\
\textbf{BridgeDPI} & \textbf{0.995}& \textbf{0.980}& \textbf{0.965}  &\textbf{0.972}\\
\hline
\multicolumn{5}{c}{human dataset} \\
\hline
k-NN&   0.860&  0.798&  0.927&  0.858\\
RF  &0.940& 0.861&  0.897&  0.879\\
L2  &0.911& 0.891&  0.913&  0.902\\
SVM &0.910& 0.966&  0.950&  0.958\\
GNN&    0.970&  0.923&  0.918&  0.920\\
E2E/GO & 0.970& 0.893& 0.914& 0.903\\
DrugVQA&    0.979&  0.954&  0.961&  0.957\\
\textbf{BridgeDPI}&\textbf{0.990}&  \textbf{0.963}& \textbf{0.949}& \textbf{0.956}\\
\hline


\end{tabular}
}
\label{tab:02}
\end{table}

\footnotetext[1]{GO feature required by E2E need to be obtained from the UniProt database, and there are a large number of GO missing for these two data sets. Therefore, we only reproduced the results of the E2E/GO (without GO feature) model. The distance map feature required by DrugVQA requires structural data from proteins. The calculation method for this part is not provided in the their code, and there are some protein deficiencies in the PDB database. As a result, we also don't offer the results of DrugVQA on the C.elegans dataset.}

\begin{figure}[h]
    \centering
    \includegraphics[trim=0 0 0 10, clip,scale=0.45]{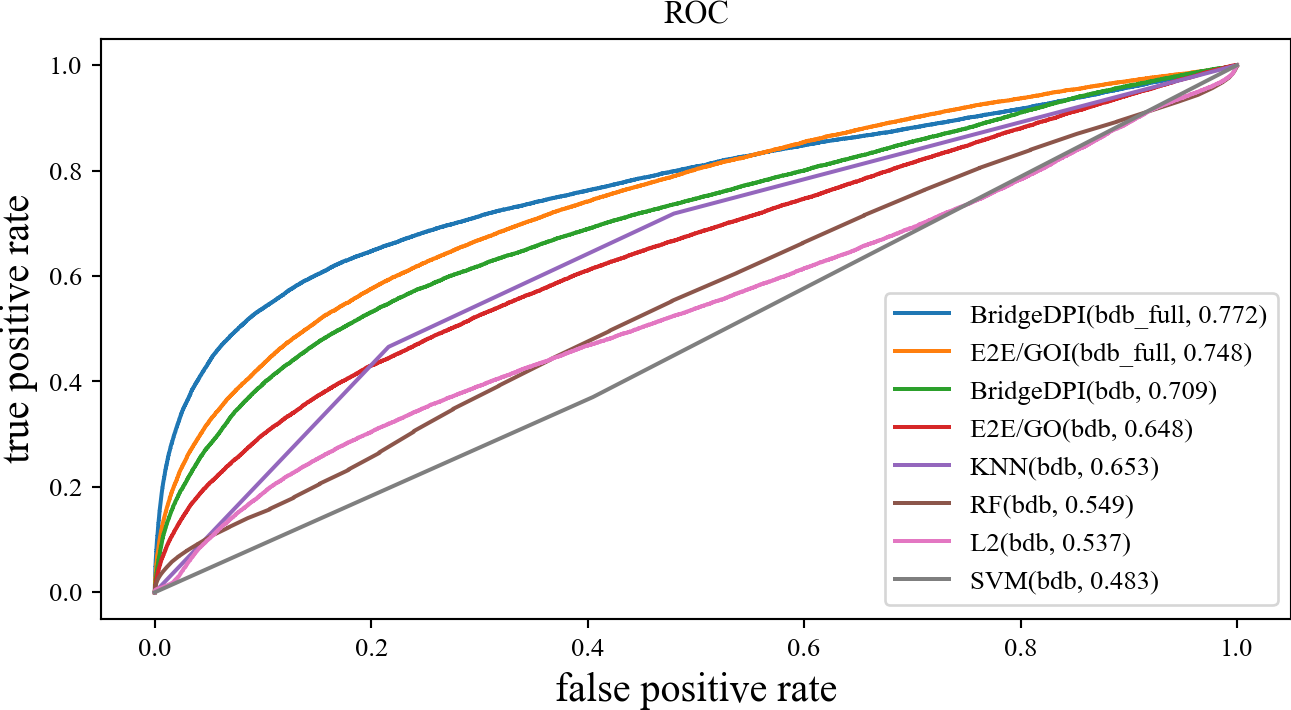}
    \caption{ \centering ROC curves of BridgeDPI with other models (training on BindingDB and testing on DUD-E)}
    \label{fig:05}
\end{figure}

Although we have achieved excellent results on these benchmark datasets, such datasets have serious data bias, which will lead to the inflated performances \citep{2019Hidden,yang2020predicting}. In order to verify the realistic performances of our model, we conduct the following experiments: train models on the customized BindingDB dataset and test models on the DUD-E dataset. We conduct 5-fold cross-validation experiments on the customized BindingDB dataset to obtain 5 models, and then predict on the DUD-E dataset to obtain 5 results, last the 5 results are averaged to evaluate the performances. The results are shown in Figure \ref{fig:05}. Similar to the previous reasons, the results for DrugVQA and E2E are not obtained. Not surprisingly, the performances of these models are greatly reduced, with AUC of the SVM even less than 0.5. Compared with other models, the AUC (0.709) of BridgeDPI is the best, which is 9.41\%, 8.58\%, 29.14\%, 32.03\%, 46.79\% higher than E2E/GO, KNN, RF, L2, SVM, respectively. Moreover, if the whole BindingDB dataset is used for training, the AUC of BridgeDPI and E2E/GO will reach to 0.772 and 0.748. The results show the effectiveness of hyper-nodes and that BridgeDPI performs better even under more realistic conditions.

\subsection{Ablative analysis}

\begin{figure}[h]
    \centering
    \includegraphics[trim=0 0 0 0, clip,scale=0.3]{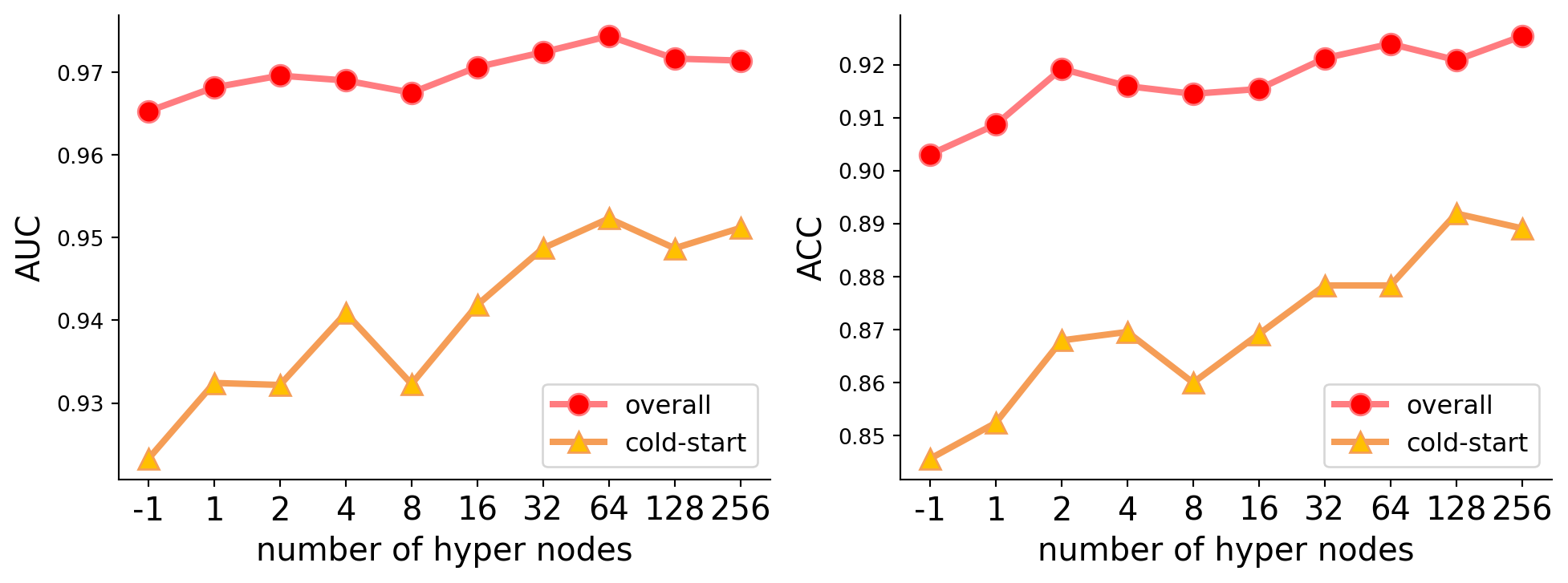}
    \caption{ \centering Performances of BridgeDPI with different number of hyper-nodes}
    \label{fig:06}
\end{figure}

By introducing the hyper-nodes, BridgeDPI builds many bridges between all proteins/drugs to conduct the prediction of DPI. In order to discover the role of the hyper-nodes, we carry out a further ablation study on the customized BindingDB dataset about the influence of the number of hyper-nodes in our model. We set the number of hyper-nodes to -1,1,2,4,8,16,32,64,128,256 separately to observe the results of our model on the test set. Among them, -1 means to remove the hyper-nodes from BridgeDTI, and the extracted feature vectors of proteins and drugs are multiplied directly and fed to the final fully connect layer. The overall AUC/ACC and the unseen proteins' AUC/ACC are mainly focused, as shown in Figure \ref{fig:06}. We can see that the introduction of the hyper-nodes can indeed improve the prediction performance of DTI, either in the overall AUC and ACC or in the unseen proteins' AUC and ACC. As the number of hyper-nodes increases, the performances are further improved. When the number is at about 64, the overall AUC and unseen proteins' AUC have reached the best values, 0.973 and 0.951 respectively. We speculate that more hyper-nodes mean more bridges between proteins/drugs. The more hyper-nodes measure the relationship between proteins/drugs together and play a similar role of voting. However, too many hyper-nodes will bring excessive costs and the risk of over-fitting. Therefore, we end up introducing 64 hyper-nodes into BridgeDPI. 

\subsection{Case study}

The outbreak of COVID-19 has caused untold damage to human society, and scientists are working hard on drug discovery. The gene sequence of COVID-19 has already been detected. In order to verify the effectiveness of BridgeDPI in practical problems, we test a variety of interactions between current possible antiviral drugs and the protein targets translated from COVID-19's viral viral gene fragments.

\begin{figure}[h]
    \centering
    \includegraphics[trim=50 0 30 0, clip,scale=0.6]{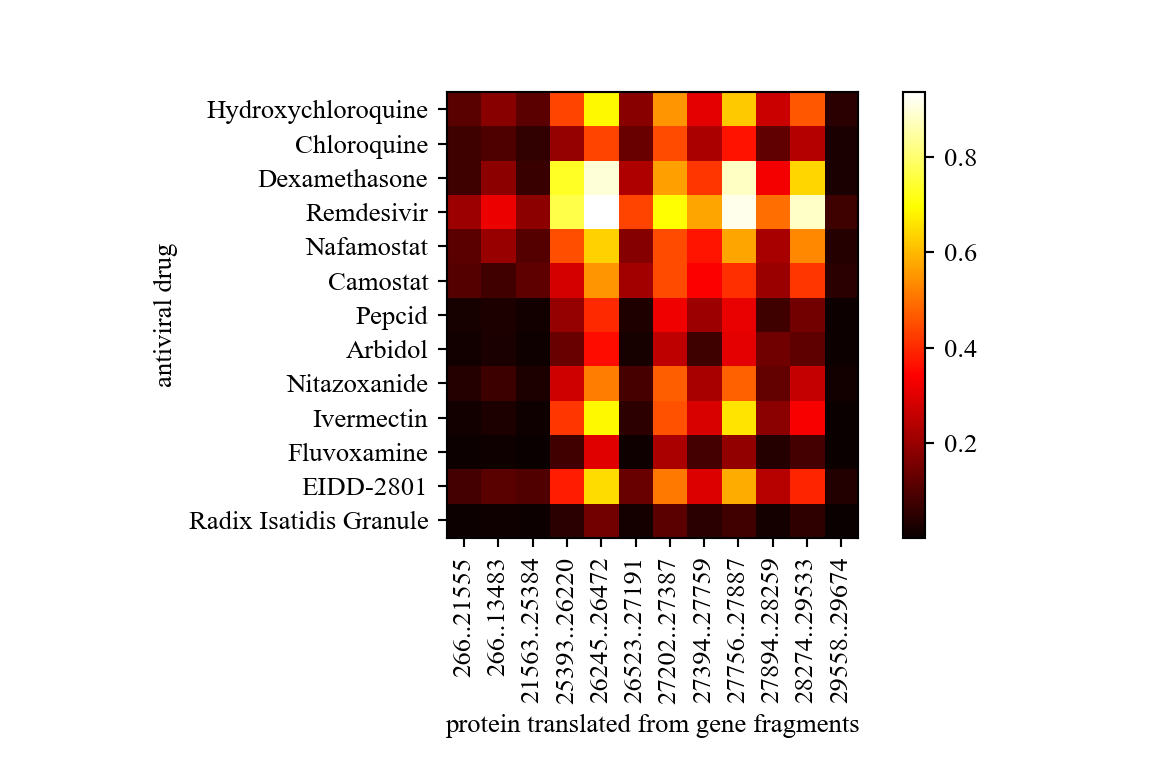}
    \caption{ \centering Predictions of antiviral drug interactions with COVID-19 virus proteins. The rows of the heatmap represent the names of antiviral drugs, the columns of the heatmap represent the gene fragments}
    \label{fig:07}
\end{figure}

First, we obtain the protein sequences translated from the COVID-19's gene fragments via NCBI database \citep{pruitt2007ncbi}. Then, these virus protein sequences and hydroxychloroquine, Chloroquine, etc. drugs are fed into BridgeDPI. Finally, the prediction results are visualized by the heat map. As can be seen from Figure \ref{fig:07}, the main potential targets of these drugs are concentrated in the protein products translated from COVID-19 gene fragment 25393 to 29533, including protein ORF3a, envelope protein, protein ORF6, protein ORF7a, protein ORF7b, protein ORF8, nucleocapsid phosphoprotein. In our results, we find that Dexamethasone and Remdesivir have the most significant effects, with the possibility of their interactions with viral protein products ORF3a, Envelope protein, ORF7b, and Nucleocapsid phosphoprotein exceeding 60\%. In fact, many studies and clinical trials have shown that the two drugs are very effective in treating COVID-19: Dexamethasone can significantly reduce mortality in COVID-19 patients \citep{2020Dexamethasone,2020Covid,2020Effect}; Remdsivir can block the replication of COVID-19 \citep{2020RemdesivirG}, reduce recovery time \citep{2020RemdesivirB} and improve survival rate \citep{2020Compassionate} in COVID-19 patients. In contrast, unrelated drugs such as Radix Isatidis Granule have little interaction potential with viral protein products. These experimental results have verified the validity and reliability of our model in predicting new drugs, indicating that BridgeDPI has a certain guiding role in the actual research and drug discovery.

\section{Conclusion}

In this work, we propose an end-to-end deep learning model to predict DPIs by introducing hyper-nodes. Hyper-nodes establish a bridge between proteins/drugs so that the information of PPA and DDA can be captured, and then get the prediction of DTIs. The experiments show that our approach outperforms other competing methods on the customized BindingDB, C.elegans, Human, DUD-E datasets and achieves state-of-the-art performances. In order to verify the realistic performances of our model, we perform cross-validation experiments on the different datasets (training on BindingDB dataset, testing on DUDE dataset) and achieve a great result. Finally, the case study with concrete examples reaffirms the usefulness of our model.

\section*{Acknowledgements}
\section*{Funding}
This work is supported in part by the NSFC-Zhejiang Joint Fundfor the Integration of Industrialization and Informatization underGrant No. U1909208, Hunan Provincial Science and TechnologyProgram 2019CB1007. 

\bibliographystyle{natbib}
\bibliography{Document}

\end{document}


\subsection{supplementary materials}

\begin{figure}[h]
    \centering
    \includegraphics[trim=50 0 50 0, clip,scale=0.3]{figure/fig5.png}
    \caption{ \centering Performances on DUD-E with training on BindingDB}
    \label{fig:05}
\end{figure}
\begin{table}[h]
    \centering
    \begin{tabular}{c|c|c|c|c}
        \hline
        Models & AUC & Precision & Recall & F1  \\
        \hline
        \multicolumn{5}{c}{C.elegans dataset} \\
        \hline
       k-NN & 0.858	&0.801	&0.827	&0.814 \\
    RF & 0.902&	0.821&	0.844&	0.832\\
    L2 & 0.892&	0.890&	0.877&	0.883\\
    SVM & 0.894	&0.785&	0.818&	0.801\\
    GNN & 0.978&0.938&	0.929&	0.933\\
   \textbf{h-GCN} & \textbf{0.995}&	\textbf{0.980}&	\textbf{0.965}	&\textbf{0.972}\\
    \hline
    \multicolumn{5}{c}{human dataset} \\
    \hline
    k-NN&	0.860&	0.798&	0.927&	0.858\\
    RF	&0.940&	0.861&	0.897&	0.879\\
    L2	&0.911&	0.891&	0.913&	0.902\\
    SVM	&0.910&	0.966&	0.950&	0.958\\
    GNN&	0.970&	0.923&	0.918&	0.920\\
    DrugVQA&	0.979&	0.954&	0.961&	0.957\\
    \textbf{h-GCN}&\textbf{0.990}&	\textbf{0.963}&	\textbf{0.949}&	\textbf{0.956}\\
    \hline
    \multicolumn{5}{c}{DUD-E dataset}\\
    \hline
    RF&	0.622&	-	&-	&-\\
    GNN&	0.940&	-	&-&	-\\
    DrugVQA	& 0.972&	-&	-&	-\\
    \textbf{h-GCN}	& \textbf{0.996}&	\textbf{0.748}&	\textbf{0.940}&	\textbf{0.831}\\
    \hline

    \end{tabular}
    \caption{ \centering Performances on the other dataset}
    \label{tab:03}
\end{table}

\begin{table*}
    \centering
    \setlength{\tabcolsep}{5mm}{
    \begin{tabular}{c|c|c|c|c|c|c}
        \hline
        \multicolumn{2}{c}{Protein feature input}&
        \multicolumn{2}{|c|}{Drug feature input}&
        \multirow{2}{*}{AUC} &\multirow{2}{*}{ ACC}& \multirow{2}{*}{F1} \\
        \cline{1-4}
  k-mers & sequence &fingerprint&sequence &\multirow{}{}{} &\multirow{}{}{} &\multirow{}{}{}  \\
        \hline
        \multicolumn{7}{c}{BindingDB dataset} \\
        \hline
       $\surd$&	&	$\surd$&	&	0.972&	0.922&	0.918\\ \hline
   $\surd$&  &   &	$\surd$&	0.961&	0.906&	0.905\\\hline
    	&$\surd$&	$\surd$&	&	0.968&	0.908&	0.907\\\hline
    &	$\surd$&		&$\surd$&	0.950&	0.879&	0.874\\\hline
    $\surd$&	$\surd$&	$\surd$&	$\surd$&	\textbf{0.974}&\textbf{	0.924}&\textbf{	0.919}\\\hline
    
   \multicolumn{7}{c}{C.elegans dataset}  \\ \hline
     $\surd$&	&	$\surd$&	&	0.989&	0.947&	0.945  \\ \hline
    $\surd$&  &   &	$\surd$&0.994&	0.967&	0.967\\ \hline
    	&$\surd$&	$\surd$&	&0.988&	0.952&	0.951\\ \hline
    &	$\surd$&		&$\surd$&	0.994&	0.968&	0.968\\ \hline
    $\surd$&	$\surd$&	$\surd$&	$\surd$&	\textbf{0.995}&	\textbf{0.971}&	\textbf{0.971}\\ \hline
    \multicolumn{7}{c}{Human dataset} \\ \hline
    $\surd$&	&	$\surd$&	&	0.984&	0.938&	0.937 \\ \hline
    $\surd$&  &   &	$\surd$&	0.985&	0.946&	0.945 \\ \hline
    	&$\surd$&	$\surd$&	&	0.983&	0.938&	0.938 \\ \hline
    &	$\surd$&		&$\surd$&	0.986&	0.945&	0.945 \\ \hline
   $\surd$&	$\surd$&	$\surd$&	$\surd$&	\textbf{0.990}&	\textbf{0.955}&	\textbf{0.954} \\ \hline

     \multicolumn{7}{c}{DUDE dataset} \\ \hline
    $\surd$&	&	$\surd$&	&	0.941&	0.962&	0.310 \\ \hline
    $\surd$&  &   &	$\surd$&	\textbf{0.994}&\textbf{0.988}&0.701 \\ \hline
    	&$\surd$&	$\surd$&	&	0.936&	0.963&	0.310 \\ \hline
    &	$\surd$&		&$\surd$&	\textbf{0.994}&	0.986&	\textbf{0.717} \\ \hline
   $\surd$&	$\surd$&	$\surd$&	$\surd$&	\textbf{0.994}&	0.975&	0.599 \\ \hline

    \end{tabular}}
    \caption{ \centering Performances on the other dataset}
    \label{tab:02}
\end{table*}